\documentclass[runningheads]{llncs}

\usepackage{latexsym}
\usepackage{url}
\usepackage{booktabs}
\usepackage{multirow, makecell}
\usepackage{pifont}
\usepackage{enumitem}
\usepackage{array}
\usepackage{xcolor}
\usepackage{subfig}
\usepackage{graphicx} 
\usepackage{booktabs} 
\usepackage{xparse}   
\usepackage{floatrow}

\NewDocumentCommand{\rot}{O{30} O{1em} m}{\makebox[#2][l]{\rotatebox{#1}{#3}}}%

\newcolumntype{L}[1]{>{\raggedright\let\newline\\\arraybackslash\hspace{0pt}}m{#1}}

\newfloatcommand{capbtabbox}{table}[][\FBwidth]

\newcommand{\dataset}{\texttt{MANtIS}}

\newcommand{\stack}{Stack Exchange}
\newcommand{\msdialog}{MSDialog}

\newcommand{\scs}{SCS}
\newcommand{\dstc}{DSTC-7-SS}
\newcommand{\udc}{UDC}

\newcommand{\mantisConvos}{80,324}
\newcommand{\mantisDomains}{14}
\newcommand{\mantisIntentConvos}{1,356}
\newcommand{\mantisEasy}{\dataset{}$^{\texttt{CRR10}}$}
\newcommand{\mantisHard}{\dataset{}$^{\texttt{CRR50}}$}

\begin{document}
\title{Introducing \dataset{}: a novel Multi-Domain Information Seeking Dialogues Dataset}

\titlerunning{Introducing \dataset{}}



\author{Gustavo Penha \and
Alexandru Balan\and
Claudia Hauff}
\authorrunning{G. Penha et al.}
%
\institute{TU Delft\\
\email{\{g.penha-1,c.hauff\}@tudelft.nl}, \\
\email{alexandru.balan0505@gmail.com}}

\date{}

\maketitle
\begin{abstract}
  Conversational  search  is  an  approach  to  information  retrieval (IR), where users engage in a dialogue with an agent in order to satisfy their information needs. Previous conceptual work described properties and actions a good agent should exhibit. Unlike them, we present a novel conceptual model defined in terms of \emph{conversational goals}, which enables us to reason about current research practices in conversational search. Based on the literature, we elicit how existing tasks and test collections from the fields of IR, natural language processing (NLP) and dialogue systems (DS) fit into this model. We describe a set of characteristics that an \emph{ideal} conversational search dataset should have. Lastly, we introduce \dataset{}\footnote{The code and dataset are available at \texttt{\url{https://guzpenha.github.io/MANtIS/}}}, a large-scale dataset containing \textbf{multi-domain} and \textbf{grounded} information seeking dialogues that fulfill all of our dataset desiderata. We provide baseline results for the conversation response ranking and user intent prediction tasks.
\end{abstract}

\section{Introduction}
\thispagestyle{plain}
\emph{Conversational search} is concerned with creating agents that fulfill an information need by means of a \emph{mixed-initiative} conversation, rather than the traditional turn-taking interaction models exhibited in search engine's results page. It is an active area of research (as evident for instance in the recent CAIR\footnote{\url{https://sites.google.com/view/cair-ws/home}} and SCAI\footnote{\url{https://scai.info/}} workshop series) due to the widespread deployment of voice-based agents, such as Google Assistant and Microsoft Cortana, and their current ineffectiveness in conducting \emph{complex} and \emph{exploratory} information seeking conversations.

Ideally, a Conversational Search System (CSS) exhibits the following competences through natural language interactions with its users~\cite{radlinski2017theoretical,azzopardi2018conceptualizing}: the CSS is able to extract, understand, refine, clarify and elicit the user information need; the CSS is able to provide answers, suggestions, summaries, recommendations, explanations, reasoning and divide the problem into sub-problems, based on its knowledge source(s); the CSS is able to take initiative, ask questions back and decide which types of actions are best suited in the current conversation context.


Current neural conversational approaches are not yet able to demonstrate these properties~\cite{gao2018neural}, as, among others, we do not have large scale and reusable training datasets that display all of the competences listed above in multi-domain information seeking conversations. At the same time, the fields of IR, NLP and DS have already engaged in relevant and intersecting sub-problems of conversational search such as 
ranking clarification questions~\cite{rao2018learning,Aliannejadi:2019:ACQ:3331184.3331265}, user intent prediction~\cite{qu2019user}, belief state tracking~\cite{chen2017survey} and conversation response ranking~\cite{yang2018response} and generation~\cite{yang2019hybrid}. 
Despite this progress, significant challenges towards building and evaluating the CSS pipeline remain. As discussed in the 2018 SWIRL report on research frontiers in IR~\cite{culpepper2018research}, two major challenges facing CSS are (1) the adaptation and aggregation of existing techniques and subsystems in IR/NLP/DS in one complex system for multi-domain information seeking dialogues and (2) the design and implementation of evaluation regimes coupled with large-scale datasets containing information seeking conversation that enable us to evaluate all desired competences of a CSS.

In this paper, we study those two challenges more closely. To deal with the first challenge we formalize a novel conceptual model, called \emph{conversational search goals}, and determine what goals of an information seeking conversation current IR/NLP/DS tasks could help achieving. Regarding the second challenge, we describe which competences of a CSS currently existing datasets are able to evaluate. We find none of the twelve recently (within the past five years) introduced datasets that we investigate to fulfill all seven of our dataset desiderata (which are described in more detail in \S{}~\ref{sec:desiredata}): multi-turn; multi-intent utterances; clarification questions; information needs; utterance labels; multi-domain; grounded. We introduce \dataset{}, a large-scale dataset that fulfills all seven of our dataset desiderata, with 80K conversations across 14 domains that we extracted from \stack{}, one of the largest question-answering portals. Lastly, we provide provide baselines for the tasks of conversation response ranking and user intent prediction.

\section{Related Work}
\begin{table}[]
\tiny
\scriptsize
\caption{Actions described by previous work on CSS. We divide them by the goals from our conceptual model.}
\label{table:related}
\begin{tabular}{@{}L{2cm}L{4.8cm}L{5cm}@{}}
\toprule
\textbf{Model} & \textbf{S1 - Information-need elucidation} & \textbf{S2 -  Information presentation}  \\ \midrule
Vakulenko et al.~\cite{vakulenko2018qrfa} & inf., understand, pos/neg feedback& prompt, offer, results, backchannel, pos/neg feedback \\
Qu et al.~\cite{qu2018analyzing} & original question, follow up question, repeat question, clarifying question, inf. request, pos/neg feedback & potential answer, further details, inf. request, pos/neg feedback  \\
Trippas et al.~\cite{trippas2017people} & query refinement offer, query repeat, query embellishment, intent clarification, confirms, inf. request& presentation, presentation with modification, presentation with modification and suggestion, scanning document, SERP, confirms, inf. request  \\ \midrule
Radlinski and Craswell~\cite{radlinski2017theoretical} &  rating of (partial) item, preference among (partial) item, lack of preference, critique of (partial) item, unstructured text describing inf. need & free text, single/partial item/cluster, small \# of partial items, small \# of partial items, complete item, small \# of complete items\\
Azzopardi et al.~\cite{azzopardi2018conceptualizing} & (non) disclose, revise, refine, expand, extract, elicit, clarify, hypothesize, interrupt & list, summarize, compare, subset, similar, repeat, back, more, note, record, recommend, report, reason, understand, explain, interrupt \\\bottomrule
\end{tabular}
\end{table}


Existing efforts in conversational search have started in late 1970's, with a dialogue-based approach for reference retrieval \cite{oddy1977information}. Since then, research in IR has focused on strategies---such as exploiting relevance feedback \cite{rocchio1971relevance}, query suggestions \cite{cao2008context} and exploratory search \cite{white2009exploratory,marchionini2006exploratory}---to make the search engine result page more interactive, which can be considered as a very crude approach to CSS. Recently, the widespread use of voice-based agents and advances in machine learning have reignited the research interest in the area. User studies~\cite{vtyurina2017exploring,thomas2017misc} have been conducted to understand how people interact with agents (simulated by humans) and inform the design of CSSs.

A number of works have defined models derived from the annotation process of collected conversational data---first three rows of Table~\ref{table:related}. Each scheme enumerates the possible user intent(s) for each utterance in the dialogue. Trippas et al.\cite{trippas2017people}~analyzed the behaviour of speech-only conversations for search tasks and defined an annotation scheme to model such interactions, which they subsequently employed to discuss search behaviour related to the type of modality (voice or text) and to the search process~\cite{trippas2018informing}. Qu et al.\cite{qu2018analyzing}~extracted information-seeking dialogues from a forum on Microsoft products to analyze user intent, using a forum annotation scheme. Vakulenko et al.\cite{vakulenko2018qrfa}~proposed a more coarse-grained model for information seeking dialogues, and based on their annotation scheme they label and analyze four different datasets via process mining. The different annotation schemes were used to get a better understanding of different aspects of the information-seeking process through dialogue.

In contrast to models derived from actual conversations, conceptual works have focused on the larger picture of CSS: theorizing about desired actions, properties and utility a CSS could have in the future, last two lines of Table~\ref{table:related}. Radlinski and Craswell~\cite{radlinski2017theoretical}~defined a framework with five desirable properties including mixed-initiative (both the user and the system can take initiative) and user-revealment (the system should help the user express and discover her information need). Additionally, they proposed a theoretical conversational search information model that exhibits such characteristics through a set of user and agent actions (e.g., displaying partial/complete items/clusters and providing feedback). This theoretical model was expanded by Azzopardi et al.~\cite{azzopardi2018conceptualizing}, who describe a set of twenty-five actions regarding possible interactions between the user and the agent, e.g., a user can \emph{revise} or \emph{refine} a criteria of her current information need; they discuss possible trade-offs between actions, highlighting future decisions and tasks for CSSs.

As pointed out by Azzopardi et al.~\cite{azzopardi2018conceptualizing}, it has not yet been discussed nor specified how to \emph{implement} the actions or decisions the agents need to perform in a CSS, thus we still need a practical way to advance the field in this direction. In order to understand how different research fields have worked with conversational search in practical terms, we define a novel model to describe information-seeking conversations, by defining the main goals of such conversations. With this model in mind, we describe a set of characteristics a conversational search dataset should have, analyze which features existing datasets have and finally introduce \dataset{}.
\section{Conversational Search Goals}

Unlike previous models~\cite{vakulenko2018qrfa,qu2018analyzing,trippas2017people,radlinski2017theoretical,azzopardi2018conceptualizing} that focus on annotation schemes and desired properties/actions of CSSs, our main objective is to understand how different research fields have tackled areas of conversational search in terms of tasks and datasets. To this end, we define a conceptual model that describes the main \emph{goals} of information seeking conversations. We opted for a model on the goal-level as it enables us to understand to what extent we can rely on existing tasks and datasets to train and evaluate CSSs.

  \begin{figure}[]
    \centering
      \includegraphics[width=0.9\textwidth]{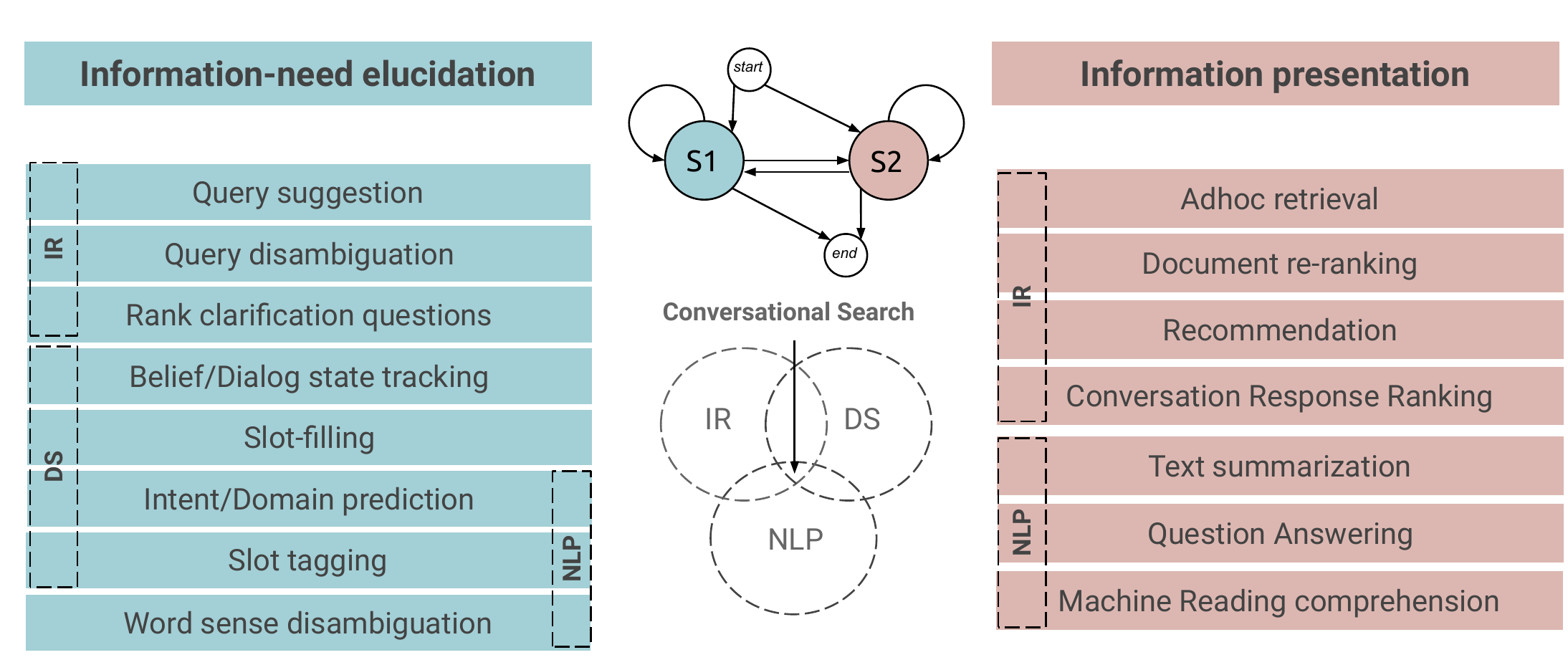}
      \caption{
      Overview of our conversational search goals model and related tasks.}
      \label{fig:conversational_pipeline}
    \end{figure}

Figure~\ref{fig:conversational_pipeline} depicts our conversational search goals model. First, we define two states in a search conversation: \emph{information-need elucidation} (S1) and \emph{information presentation} (S2). We believe them to be the two major goals pursued by the agent during the progression of information-seeking dialogues. Arrows indicate either user or agent utterances during the conversation, which might lead to a transition between goals or development under the same goals. Let us now describe the goals from our model and connect tasks from the related research fields to them. 



\subsubsection*{State 1: Information-need elucidation} \label{sec:inf_need_elucidation}

An important role of a CSS is helping the user understand, clarify, refine, express and elicit their information need~\cite{azzopardi2018conceptualizing}; this is one key difference from traditional search engines~\cite{culpepper2018research}. The IR, NLP and DS communities offer only partial perspectives into this goal. From the IR point of view, this challenge has been tackled with query suggestions and query disambiguation techniques. Such methods are trained and evaluated using search engine query logs, which are not mixed-initiative nor dialogue-based and hence not sufficient for training and testing CSSs' capabilities of elucidating information needs.

The DS community has focused on representing the user information-need with explicit pre-defined slots and values that are extracted from user utterances, and accumulated as a belief state. This approach is not directly applicable to CSS, as it is not viable to enumerate all possible slots/values combinations for open-domain information-seeking dialogues. 

Related work in NLP includes predicting the intent or domain of each utterance~\cite{qu2019user}, and learning representations of the user information need through its context (previous utterances) \cite{wu2017sequential,kenter2017attentive} in order to complete a downstream task, e.g. response generation. Another relevant task that relates to both NLP and IR is using information-seeking datasets extracted from online forums, e.g. \stack~\cite{rao2018learning} and \msdialog{}~\cite{qu2018analyzing}, to rank/generate clarification questions given the dialogue context. 

\subsubsection*{State 2: Information presentation}
The other conversational goal is to extract/retrieve and present the \emph{relevant} information in a conversational manner. The system has to decide \emph{how} and \emph{which} information to present. In this stage of the conversation, the agent provides answers, suggestions, summaries, explanations, recommendations, reasoning and possibly divides the problem into sub-problems, all based on its knowledge sources, e.g. document corpora, databases or sets of existing user answers from online fora. The user is in charge of evaluating and making sense of the presented information, giving feedback and asking for further information.

In IR, approaches have taken into account the previous queries and implicit user feedback in search sessions, such as clicks on documents and dwell time, which can be useful resources for the search engine to retrieve the next batch of results in the search session~\cite{joachims2005accurately,guo2012beyond}. In IR and NLP there is extensive research focused on developing models that improve web retrieval and question-answering regardless of the user interactions with the system. Related tasks include ad-hoc retrieval, document re-ranking, recommendation, machine reading comprehension, answer generation/ranking, and text summarization among others. The main open challenge here is evaluating and adapting extraction and presentation techniques for information-seeking dialogues.

From the traditional DS perspective, this problem is delegated to the last component of the system's pipeline\footnote{A generic dialogue system is composed of the following: natural language understanding $\rightarrow$ dialogue state tracking $\rightarrow$ policy learning $\rightarrow$ natural language generation \cite{chen2017survey}.} where natural language generation is used to deliver the response based on the state of the dialogue. This requires the system to determine which action to take at the moment as a natural language response, given its knowledge source and the current belief state. In order to translate the traditional DS approach to CSSs, we need to know in advance the domain of the conversation, i.e., employ a domain classifier, and manually engineer a suitable system for each domain, which is difficult to scale for open-domain information needs.

\subsubsection*{States transitions}
During the dialogue, the CSS can choose between a number of actions; it has to decide which one(s) to take and then provide a natural language response to the user. Learning a mapping between the next action based on the current conversation state has been evaluated in the DS community through the task of dialog policy learning \cite{peng2018deep,tang2018subgoal}. In goal-oriented dialogues we can manually define a set of domain-dependent actions, e.g., compare products and recommend. NLP generally handles this with distributed representations of dialogues and information needs, which are learned in a end-to-end manner to generate answers~\cite{gao2018neural}. One of the challenges in conversational search is for the system to determine when to move between the goals of the conversation. CSSs can have mechanisms that handle this explicitly or do it in a fully data-driven and end-to-end manner.

\section{Dataset Desiderata} \label{sec:desiredata}



Despite the fact that the IR, NLP and DS communities have independently contributed to aspects of conversational search, we argue that we currently cannot fully train and evaluate the effectiveness of CSSs with existing datasets. Based on the existing theoretical frameworks of CSSs~\cite{radlinski2017theoretical,azzopardi2018conceptualizing} and our conversational search goals model we formally define a dataset desiderata:

\begin{itemize}[leftmargin=*]
    \item \textbf{Multi-turn dialogues}: the data must contain dialogues with more than one turn of user and agent utterances. Single-turn dialogues do not take into account the process of elucidating the user information-need.

    \item \textbf{Information needs}: the user must have an information need~\cite{taylor1962process} expressed in her utterances. The conversations must be information-seeking, going beyond lookup, chit-chat and goal-oriented tasks. Conversational \emph{search} is different from general conversational \emph{AI}~\cite{gao2018neural}, as there is an underlying information need to be solved.

    \item \textbf{Clarification questions}: the data must present mixed-initiative conversations by going beyond the user-asks/system-responds loop. Clarification questions are essential in elucidating the user information-need. 

    \item \textbf{Multi-intent utterances}: another indication of mixed-initiative \cite{radlinski2017theoretical} are utterances that have more than one intent such as giving positive feedback and presenting further information.
    Having \textbf{utterance labels} is a useful resource in building CSSs 
    by providing additional supervision signals.
    
    \item \textbf{Multi-domain}: the users' information needs can fall into more than one domain (topics of conversation, such as \textit{physics}, \textit{travel} and \textit{English}). Domain specific dialogue systems do not generalize to new/unseen information needs. Thus the dataset must contain conversations from multiple domains.
    
    \item \textbf{Grounded conversations}: the agent must be able to report the source(s) of the information it is providing and the reasoning behind it. Grounding conversations in documents is a useful resource for achieving explainable agents.
    Moreover, using sources of information for generating responses has shown to improve the quality of the dialogues over non-grounded conversations that rely only on historical conversational data \cite{zhou2018dataset}.
    
\end{itemize}

\begin{table*}[]
\centering
\caption{Overview of dialogue datasets including their size and conversational search characteristics. \footnotesize{$^{a}$ \textit{The dialog acts were pre-defined, and the teacher in the setup chooses only one among few options.} $^b$ \textit{There are labels for a sample of 2,199 dialogues.} $^c$ \textit{There are labels for a sample of \mantisIntentConvos{} dialogues.}}}
\label{table:datasets}
\begin{tabular}{@{}lllrlllllll@{}}
\textbf{Name}                                                 & \textbf{Venue} & \textbf{Field} & \textbf{\#Dialogues} & \rot{\textbf{multi-turn}} & \rot{\textbf{multi-intent}} & \rot{\textbf{clf. questions}} & \rot{\textbf{inf. needs}} & \rot{\textbf{utterance labels}} & \rot{\textbf{multi-domain}} & \rot{\textbf{grounded}} \\ \midrule
SCS  \cite{trippas2018informing,trippas2017people}            & CHIIR               & IR                      & 39                       & \ding{51}                 & \ding{51}                              & \ding{51}                              & \ding{51}                                & \ding{51}                        & \ding{51}                  &                                       \\
MISC \cite{thomas2017misc}                                    & CAIR workshop               & IR                      & 88                       & \ding{51}                 & \ding{51}                              & \ding{51}                              & \ding{51}                                &                                  & \ding{51}                  &                                       \\
CCPE-M \cite{48414}                                    & SIGDIAL               & DS                      & 502                       & \ding{51}                 & \ding{51}                              & \ding{51}                              &                                 &    \ding{51}                              & \ding{51}                  &                                       \\
Frames \cite{asri2017frames}                                  & SIGDIAL             & DS                      & 1,369                     & \ding{51}                 & \ding{51}                              & \ding{51}                              &                                          & \ding{51}                        &                            &                                       \\
KVRET \cite{eric2017key}                                      & SIGDIAL             & DS                      & 3,031                     & \ding{51}                 & \ding{51}                              &                                        &                                          & \ding{51}                        &                            & \ding{51}                             \\
CoQA \cite{reddy2018coqa}                                     & preprint only       &             -            & 8,000                     & \ding{51}                 &                                        &                                        &                                          &                                  & \ding{51}                  & \ding{51}                             \\
MultiWOZ \cite{budzianowski2018multiwoz}                      & EMNLP               & NLP                     & 8,438                     & \ding{51}                 & \ding{51}                              & \ding{51}                              &                                          & \ding{51}                        &                            &                                       \\
QuAC \cite{choi2018quac}                                      & EMNLP               & NLP                     & 13,594                    & \ding{51}                 &                                        &                                        &                                          & \ding{51}  $^{a}$                & \ding{51}                  & \ding{51}                             \\
WoW  \cite{dinan2018wizard}                                   & ICLR             & ML                      & 22,311                    & \ding{51}                 & \ding{51}                              &                                        &                                          &                                  & \ding{51}                  & \ding{51}                             \\
ShARC \cite{saeidi2018interpretation}                         & EMNLP               & NLP                     & 32,436                    & \ding{51}                 &                                        &                                        &                                          &                                  &                            & \ding{51}                             \\
MSDialog \cite{qu2018analyzing}                               & SIGIR               & IR                      & 35,000                    & \ding{51}                 & \ding{51}                              & \ding{51}                              & \ding{51}                                & \ding{51} $^{b}$                 &                            &                                       \\
DSTC-7-SS \cite{DSTC7} & DSTC7 workshop      & DS                      & 100,000                   & \ding{51}                 & \ding{51}                              & \ding{51}                              & \ding{51}                                &                                  &                            &                                       \\
UDC \cite{lowe2015ubuntu}                                     & SIGDIAL             & DS                      & 930,000                   & \ding{51}                 & \ding{51}                              & \ding{51}                              & \ding{51}                                &                                  &                            &                                       \\ \midrule
\textbf{\dataset{}} &   -   &     -                                                          & \mantisConvos{}                & \ding{51}                 & \ding{51}                              & \ding{51}                              & \ding{51}                                &       \ding{51} $^{c}$                          & \ding{51}                           &  \ding{51}    \\
\bottomrule
\end{tabular}
\end{table*}

With this desiderata in mind, we explored twelve multi-turn, non-chit-chat, human-to-human, open-sourced and recent (last 5 years) datasets in detail. The result---i.e. the datasets' characteristics according to our desiderata---can be found in Table~\ref{table:datasets}. Importantly, none of the datasets have all the desirable features. \scs{} is the most complete one, missing only the grounding aspect. However, the very limited number of dialogues in this dataset (39) makes it not suitable to train and evaluate conversational search models. The three largest datasets, \msdialog{}, \dstc{} and \udc{}, were all derived from technical forums. Their two main drawbacks are the narrow content domain (technical) and lack of a correspondence between utterances and documents where useful information to fulfill the information needs could be extracted from (i.e., grounding). This poses challenges for research on CSS: how generalizable are models trained on one or two particular domains? How can systems leverage the huge amount of available information in web documents---from diverse domains---in information-seeking conversations? 

In order to study such challenges we created a novel dataset called \dataset{}, short for \textit{\underline{m}ulti-dom\underline{a}i\underline{n} \underline{i}nformation \underline{s}eeking dialogues dataset}. \dataset{} is to our knowledge the first dataset at large-scale that fulfills all of our dataset desiderata.


\section{\dataset{}}

\begin{figure*}
\centering
  \includegraphics[width=0.9\textwidth]{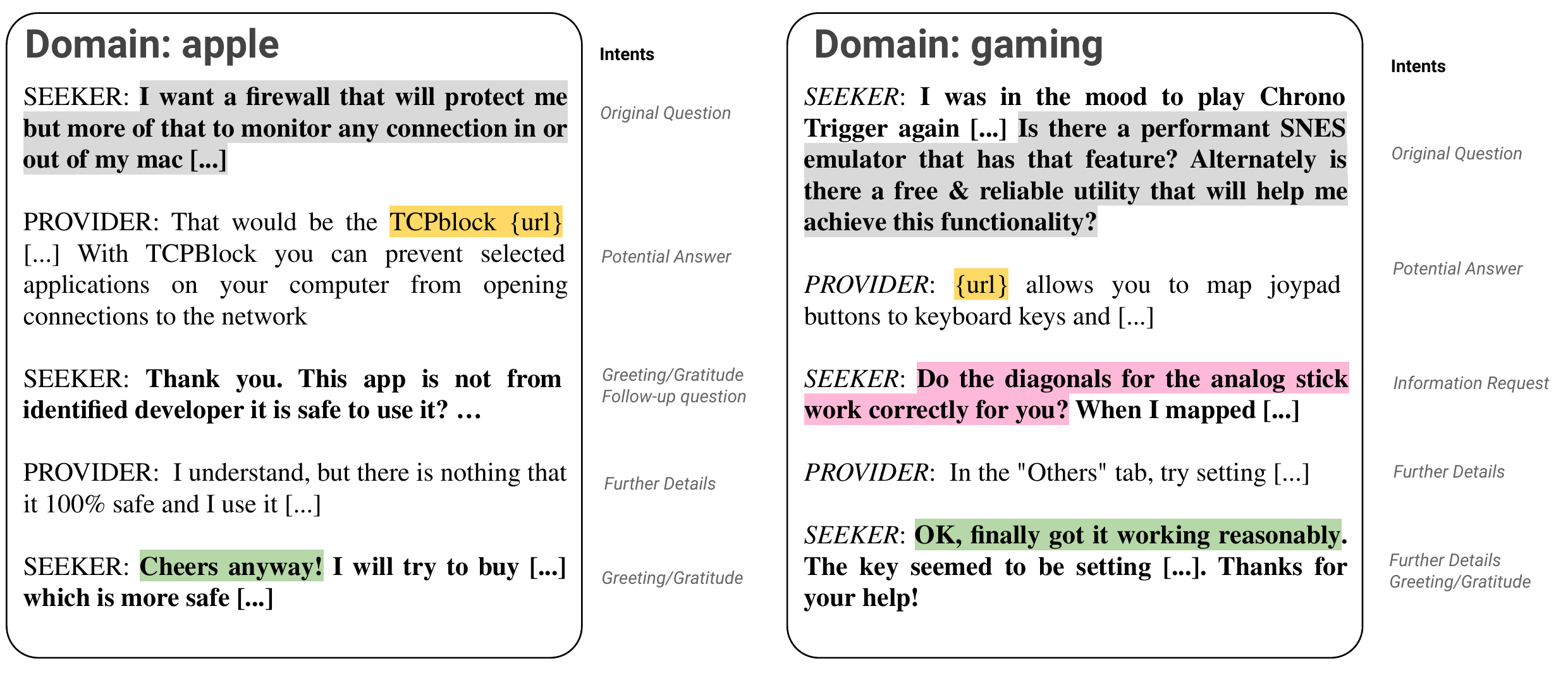}
  \caption{\dataset{} examples with document grounding (yellow), positive feedback from the information seeking user (green), clarification questions (pink) and the initial information need (gray). On the right we display the user intent labels.}
    \label{fig:sites}
\end{figure*}



\begin{figure}[ht!]
    \centering
    \subfloat[average number of turns by domain]{{\includegraphics[width=0.45\textwidth]{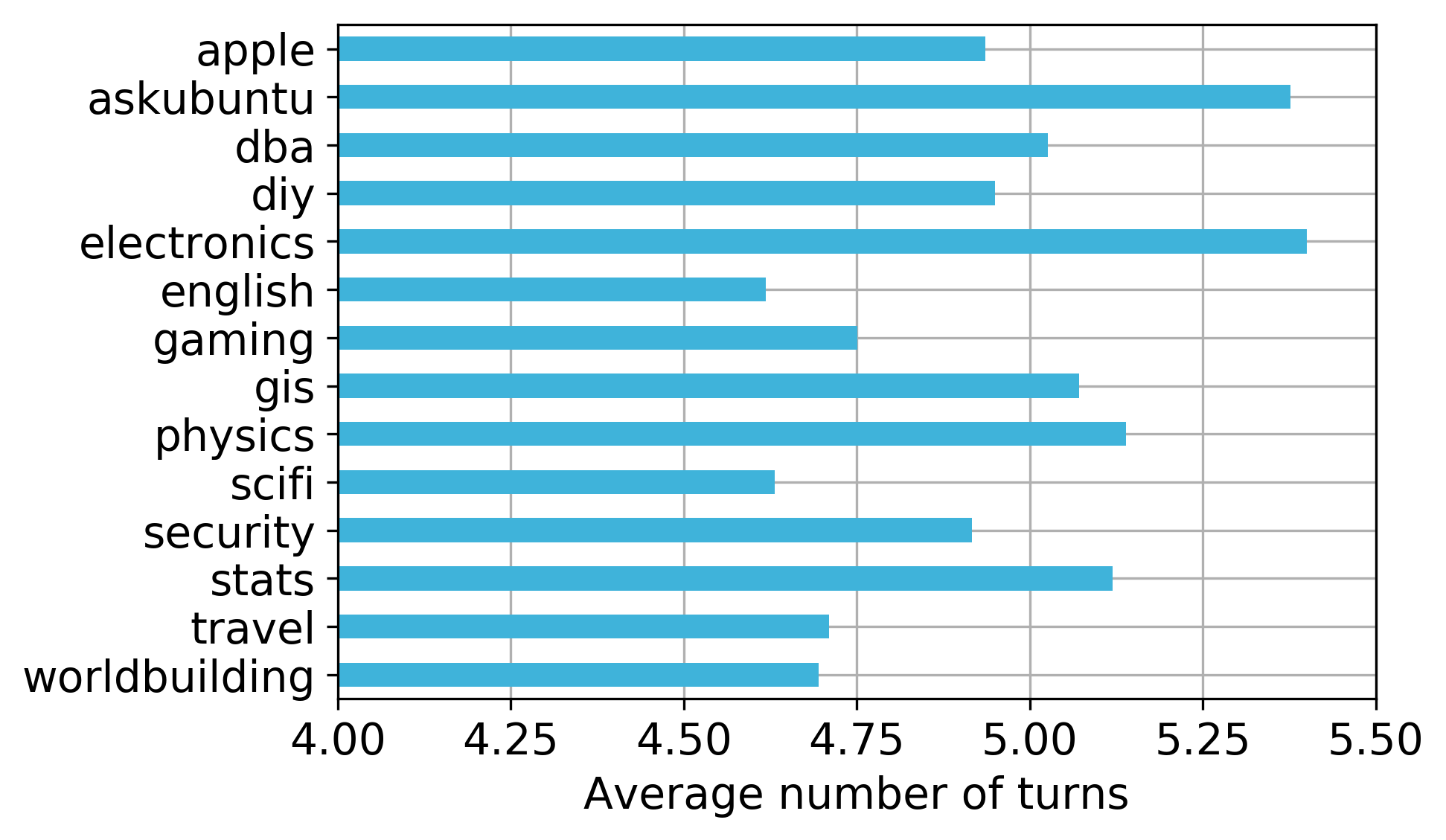} \label{fig:turns_dist}}}
    \subfloat[average number of terms by domain]{{\includegraphics[width=0.45\textwidth]{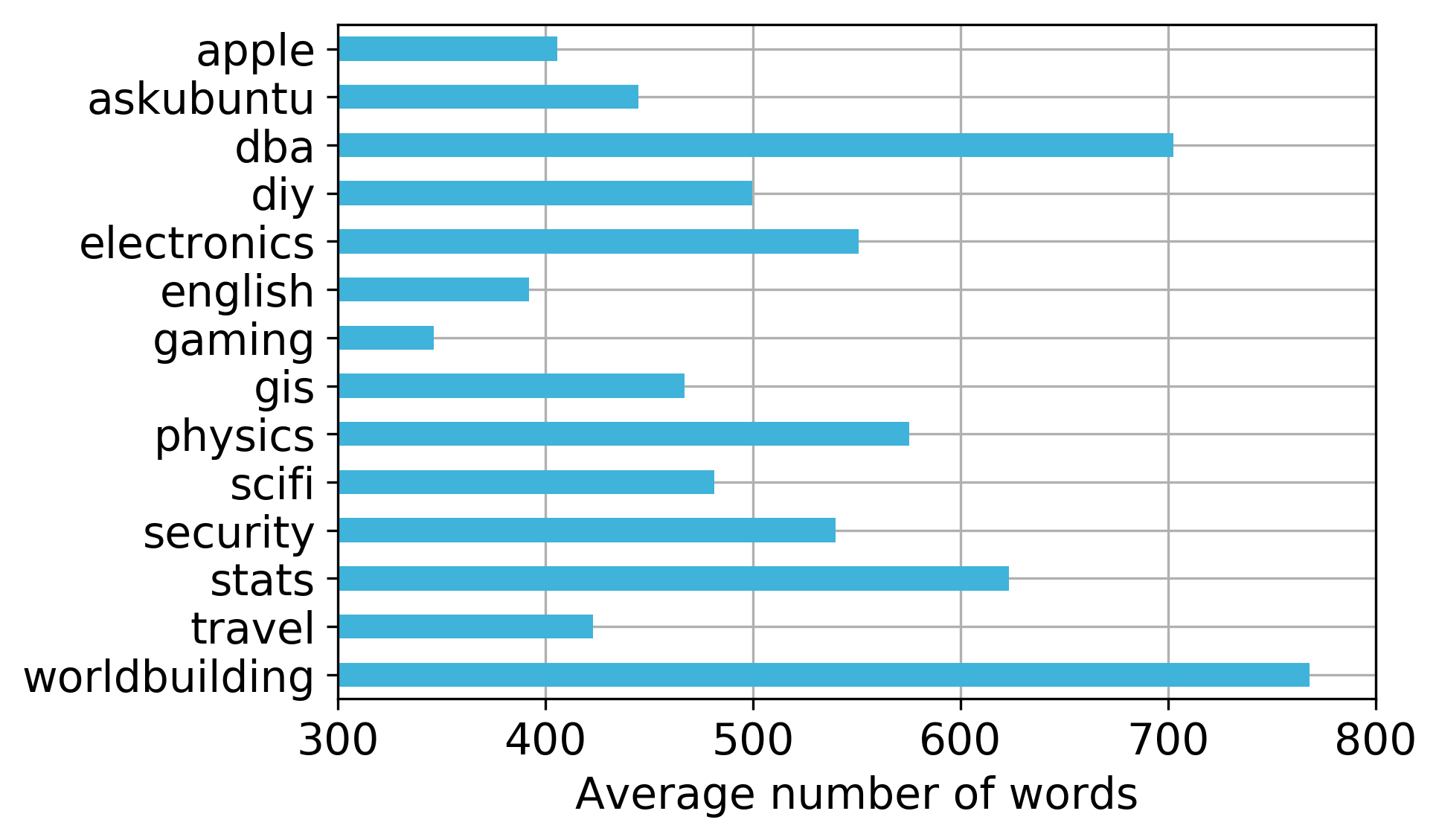} \label{fig:words_dist}}}
    \caption{\dataset{} quantitative analysis.}
    \label{fig:my_label}
\end{figure}

In order to create a large-scale conversational dataset, we resort to the extraction of conversations from existing data sources---same strategy followed by the creators of the largest datasets in Table~\ref{table:datasets}.  We take the community question-answering portal \stack{} as a starting point\footnote{\url{https://archive.org/download/stackexchange} data dump from 2019-03-04} as (i) the data dump is publicly available, (ii) it is large-scale (more than 20M questions), (iii) the portal covers diverse domains (so-called \emph{sites}, 175 as of 05/2019) such as \textit{physics}, \textit{travel} and a range of IT and computer science domains, and (iv) the information needs are often complex as posing a question on \stack{} usually means that a simple web search is not enough to find a suitable answer. 

For \dataset{}, we consider \mantisDomains{} diverse domains\footnote{Specifically, we consider \texttt{apple} (5,645 dialogues), \texttt{askubuntu} (17,755), \texttt{dba} (5,197), \texttt{diy} (1,528), \texttt{electronics}(10,690), \texttt{english} (3,231), \texttt{gaming} (2,982), \texttt{gis} (9,095), \texttt{physics} (7,826), \texttt{scifi} (2,214), \texttt{security} (3,752), \texttt{stats} (7,676), \texttt{travel} (1,433) and \texttt{worldbuilding} (1,300).}. We make the source code available at \texttt{[Anonymized for review purposes]} so that conversations from any of the 175 domains of \stack{} can be extracted. The examples in Figure~\ref{fig:sites} showcase characteristics of the conversations from our dataset.

\subsubsection*{Inclusion Criteria}

We consider each question-answering thread of a Stack Exchange site as potential conversation between an information seeker and an information provider and include it in \dataset{} if the following six criteria hold:

\begin{enumerate} 
    \item The entire conversation takes place between exactly two users (the information \emph{seeker} who starts off the conversation and the information \emph{provider}).
    \item The conversation consists of at least 2 utterances per user.
    \item One of the provider's utterances contains a hyperlink, providing grounding.
    \item The conversation has not been marked as \textit{Spam} or \textit{Offensive}.
    \item The conversation has not been edited or marked as deprecated. 
    \item If the final utterance belongs to the seeker, it contains positive feedback. 
\end{enumerate}

In order to verify to what extent the existence of a hyperlink can be considered as document grounding (criterium 3), we sampled 150 conversations from \dataset{} and manually verified whether the link contained in the information provider's utterance(s) is indeed leading to a grounding document. This was the case for 88\% of the sampled conversations, which we consider a sufficiently high percentage to not further refine the grounding rule.

In order to verify whether the final say of the information seeker was a positive statement (criterium 6), we sampled 1,400 conversations (100 from each of our sites) where the last person to respond was the information seeker and manually assessed whether the final response was positive feedback (see green highlights in Figure~\ref{fig:sites}). Subsequently, for all conversations with a final response by the information seeker we computed the VADER sentiment score~\cite{hutto2014vader}. Based on our labelled conversations, we applied a decision stump in order to obtain the optimal score threshold (separately for each site). Consequently, all the conversations with a VADER score below the optimal threshold were discarded---as we are interested in information-seeking conversations that contain a positive conclusion as we assume that in those cases the information need has been fulfilled.

Based on these criteria, we extracted a total of \mantisConvos{} conversations. The majority of the conversations have 4 utterances (60\%). Some technical domains such as \textit{electronics} and \textit{askubuntu} have high average number of turns, Figure~\ref{fig:turns_dist}, while other domains such as \textit{worldbuilding} and \textit{dba} have very long utterances, Figure \ref{fig:words_dist}, showcasing the diversity of the domain's distributions. Our list of conditions were quite stringent, only 4.77\% of all question-answering threads made it into our final dataset, each domain contributed at least 1K conversations.

\subsubsection*{Conversation Labelling}

\begin{table*}[ht]
\scriptsize
\centering
\caption{User intent annotation scheme and \dataset{} label distribution.}
\begin{tabular}{@{}llr@{}}
\toprule
\textbf{Category} & \textbf{Description} & \textbf{Percentage} \\ \midrule
Original Question & The original question posed by the seeker & 16.27\% \\
Further Details & A user provides more details. & 27.72\% \\
Follow Up Question & Seeker asks one or more follow up questions. & 5.21\% \\
Information Request & A user asking for clarifications or further information. & 10.39\% \\
Potential Answer & A potential solution, given by the information provider. & 18.63\% \\
Positive Feedback & Seeker provides positive feedback about the response. & 4.73\% \\
Negative Feedback & Seeker provides negative feedback about response. & 4.03\% \\
Greetings / Gratitude & A user offers a greeting or expresses gratitude. & 10.13\% \\
Other & Anything that does not fit into the above categories. & 2.84 \% \\ \bottomrule
\end{tabular}
\label{table:intent_schema}
\end{table*}


As previous research has shown~\cite{trippas2018informing}, agents are required to elicit more information, such as user relevance feedback or further information requests in order to give the best answer. To be able to detect these types of user intent, we sampled \mantisIntentConvos{} conversations from \dataset{} and manually labeled their utterances according to the user intent, resulting in a total of 6,701 labelled utterances. Based on \cite{qu2019user}, we have defined nine types of labels, which are either related to a question (\textit{Original question}, \textit{Follow Up Question}, \textit{Information Request}), to an answer (\textit{Potential Answer}, \textit{Further Details}), expresses gratitude (\textit{Greeting/Gratitude}) or indicate feedback (\textit{Positive Feedback}, \textit{Negative Feedback}). Any utterance that does not fall in at least one of the aforementioned categories is labeled as \textit{Other}. The full description of each intent is described at Table~\ref{table:intent_schema}.

An utterance can be annotated with more than one label (cf. Figure~\ref{fig:sites}). Two expert annotators labelled each utterance within our sampled conversations; $151$ utterances were labelled by both annotators to determine the agreement between the annotators~\cite{krippendorff2011computing}, leading to a Krippendorff's $\alpha$ of 0.71. \emph{Original Question}, \emph{Potential Answer} and \emph{Further Details} are the most frequent labels. $21\%$ of utterances were annotated with more than one label, indicating the multi-intent nature of our dataset.



\section{Tasks and Baselines}

The dataset is suitable for a range of tasks including conversation response ranking~\cite{yang2018response,lowe2015ubuntu,reddy2018coqa}, conversation response generation~\cite{yang2019hybrid} and user intent prediction~\cite{qu2018analyzing,qu2019user}. We now first discuss the conversation response ranking task and then move on to the intent prediction task. Both tasks are typically seen as a step towards conversational search.

\subsection*{Conversation Response Ranking}

The typical setup of conversation response ranking is as follows: given the conversation history and the current utterance, predict the correct reply from a set of possible replies. We provide two flavours of \dataset{} to accommodate this task, following the procedure by~\cite{yang2018response}: \mantisEasy{} and \mantisHard{}. For each conversation (with $n_a$ agent replies) in \dataset{} we create a number of \emph{conversation contexts} equal to $n_a-1$ (a context is the entire conversation history up to the agent reply), as we are not taking into account contexts that have just a single agent reply. Thus, a conversation consisting of two utterances per user will yield a single conversation context.
In total, there are 118,349 conversation contexts in \dataset{}. The reply as found in \dataset{} is considered the ground-truth reply. For each ground truth reply we also sample a set of negative replies by using the ground truth reply as query in a retrieval system setup and ranking all possible replies from the training and development set according to their BM25~\cite{Robertson94BM25} retrieval score. We then sample 10 (for \mantisEasy{}) and 50 (for \mantisHard{}) negative (i.e. non-relevant) replies respectively from the top 1K ranked replies. For \mantisEasy{}, the ranked replies originate from the same \textit{site}, while for \mantisHard{} the reply can come from any domain.  

We provide a default split for the dataset into training, development and test set based on time in a 70/15/15 fashion: the oldest 70\% of conversations are the training set, while the most recent 15\% of conversations are part of the test set.

As baselines, we provide three models: (1) \emph{BM25} (the concatenation of all the utterances in the conversation context is used as query); (2) Deep Matching Network (\emph{DMN})\footnote{\url{https://github.com/yangliuy/NeuralResponseRanking}}~\cite{wu2017sequential}, which is an interaction-focused neural ranking model that creates matching matrices between each utterance in the conversation so far and the candidate response; (3) fine-tuned \emph{BERT}~\cite{devlin2019bert} using the CLS token following~\cite{yang2019simple}; here, the concatenation of all the utterances in the conversation context is used as query which is separated from the document by a SEP token.

Results displayed at Table~\ref{table:cross_datasets_results} show unsurprisingly~\cite{yang2019simple,yang2019end} that BERT is the best performing model, with a large effectiveness increase over the strong neural baseline DMN. However, with only fifty response options the models degrade severely, indicating that in more realistic settings, with potentially hundreds of thousands of responses to choose from, current approaches fail.

\begin{table*}[ht!]
\centering
\footnotesize
\caption{Baseline test set results for the conversational response ranking task, averaged over 5 runs.}
\label{table:cross_datasets_results}
\begin{tabular}{@{}lllll@{}}
\toprule
 & \multicolumn{2}{c}{\mantisEasy{}} & \multicolumn{2}{c}{\mantisHard{}} \\ \cmidrule(lr{1em}){2-3} \cmidrule(lr{1em}){4-5} 
 & MAP & nDCG@10 & MAP & nDCG@10 \\\cmidrule(lr{1em}){2-3} \cmidrule(lr{1em}){4-5} 
BM25 & 0.317 (-) & 0.475 (-) & 0.163 (-) & 0.195 (-) \\
DMN  & 0.683 (.02) & 0.761 (.015) & 0.43 (.03) & 0.512 (.038)\\ 
BERT  &\textbf{0.733} (.003)& \textbf{0.799} (.002)& \textbf{0.519} (.005) & \textbf{0.583} (.005)\\ 
\bottomrule
\end{tabular}
\end{table*}

\subsection*{User Intent Prediction}

The setup of the user intent prediction task is as follows: predict the user intents for each utterance in a given conversation. As one utterance can have multiple intents, this is a multi-label and multi-class text classification problem, and our subset of \mantisIntentConvos{} manually labeled conversations resulting in 6,701 labeled utterances is suited for this task.

We provide non-neural baselines that represent each utterance using bag-of-words with TF-IDF term weighting, following previous work on text classification ~\cite{campos2017stacking}. We use the following learning algorithms: SVM, AdaBoost, Gradient Boosting and Logistic Regression~\footnote{We use scikit-learn implementations \url{http://scikit-learn.org/}}. We employ the \textit{One vs Rest} classification strategy in order to the reduce the problem of multi-class classification to multiple binary classification problems. 

Additionally, we provide two neural baseline models that learn textual representations. The first is a BiGRU, where each word is represented via pre-trained word embeddings (word2vec\footnote{{\url{https://github.com/dav/word2vec.git}}} and fed to a BiGRU layer to learn sentence representations. The states of all words of BiGRU are passed to a fully connected layer with a softmax activation function for making the predictions. The model is trained with the cross entropy loss, using Adam optimizer. The second one is fine-tuned BERT~\cite{devlin2019bert}, where each entry contain a single utterance and we use the CLS token to predict the intents using a fully connected layer.

 
We evaluate approaches using 10-fold cross-validation (we do not use the 70/15/15 split here due to the limited amount of labeled utterances), and report the average of : Precision, Micro and Macro F1. The results, Table \ref{table:intent_prediction_results}, show that the best performing model is BERT, with a 0.13 absolute improvement in precision over the second best-performing model, the Gradient Boosting with bag-of-words model. Regarding models that do not employ heavy pre-training, AdaBoost and Gradient Boosting perform better than the BiGRU neural architecture, consistent with previous research~\cite{qu2018analyzing}.

\begin{table}[]
\centering
\footnotesize
\begin{tabular}{llll}
\toprule
Classifier                   & Precision & F1-Micro & F1-Macro \\
\midrule
LogisticRegression & 0.486 (.017) & 0.469 (.014) & 0.348 (.014) \\
SVM & 0.532 (.021) & 0.534 (.019) & 0.455 (.018) \\
BiGRU & 0.574 (.016) & 0.563 (.015) & 0.478 (.027) \\
AdaBoost  & 0.641 (.015) & 0.585 (.012) & 0.480 (.010) \\
GradientBoosting & 0.657 (.017)     & 0.611 (.013)    & 0.491 (.011)   \\
BERT & \textbf{0.790} (.013) & \textbf{0.750} (.015) & \textbf{0.591} (.030) \\
\bottomrule
\end{tabular}
\caption{Baseline results for the user intent prediction task, average and standard deviation of the cross-validation (k=10).}
\label{table:intent_prediction_results}
\end{table}

\section{Conclusions}

Conversational search is becoming an increasingly important research topic, due to the rise of voice-enabled devices. In this work, we have proposed a model of conversational search that focuses on the main goals of the agent and user interactions. We identified two major challenges: (1) the collaboration of efforts in the research fields of IR, NLP and DS, and (2) the lack of publicly available large-scale conversational search datasets. Based on a set of dataset desiderata, we created \dataset{}, a large-scale conversational search dataset that contains more than 80K conversations across 14 domains that are multi-turn, centered around complex information needs and are mixed-initative. Finally, we explored to what extent current models are able to perform well on the tasks of conversation response ranking and intent prediction---our results indicate that there is some way to go in order to deploy them in a realistic setting.

\bibliography{main}
\bibliographystyle{splncs04}

\appendix


\end{document}